%% file: main.tex
\documentclass[10pt,letterpaper]{article}
\input{preamble}
\hypersetup{
  pdftitle={Spark-H3: Adaptive Block Sparse Attention for Video Diffusion Models},
  pdfauthor={Zecheng Tang; Jiaye Fu; Youtian Lin; Xinyao Rao; Liuxuan Sun; Jian Zhang; Yao Yao; Wei Li}
}
\newcommand{\reportdate}{November 18, 2026}
\begin{document}
\newgeometry{left=20mm,right=20mm,top=16mm,bottom=16mm}
\input{cover}
\clearpage
\restoregeometry

\input{sec/spark_1_intro}
\input{sec/spark_2_related}
\input{sec/spark_3_method}

\FloatBarrier
\input{sec/spark_4_experiments}
\FloatBarrier
\input{sec/spark_5_discussion}
\clearpage
\phantomsection
\addcontentsline{toc}{section}{References}
{\small\bibliographystyle{ieeenat_fullname}\bibliography{spark_h3}}
\end{document}

%% file: preamble.tex
\usepackage[T1]{fontenc}
\usepackage[utf8]{inputenc}
\usepackage{tgadventor}
\usepackage[letterpaper,left=25mm,right=25mm,top=25mm,bottom=25mm,
  headheight=21pt,headsep=14pt,footskip=24pt]{geometry}
\usepackage[dvipsnames]{xcolor}
\usepackage{graphicx,booktabs,colortbl,array}
\usepackage{amsmath,amssymb}
\usepackage{algorithm,algpseudocode}
\usepackage{fix-cm,placeins,flafter}
\usepackage[numbers,sort&compress]{natbib}
\usepackage{microtype,xspace,etoolbox}
\usepackage[hyphens]{url}
\usepackage{titlesec,tocloft,fancyhdr}
\usepackage[most]{tcolorbox}
\usepackage{fontawesome5}
\usepackage[font=small,labelfont=bf,labelsep=period]{caption}
\usepackage{subcaption}
\usepackage{enumitem}

\definecolor{pkured}{rgb}{0.62352941,0.00000000,0.08074482}
\definecolor{pkupale}{rgb}{1.00000000,0.94117647,0.94879391}
\definecolor{pkubrand}{RGB}{139,0,18}
\definecolor{teaserdense}{HTML}{C51C65}
\definecolor{teaserspark}{HTML}{0D7561}
\definecolor{teaserfullysparse}{HTML}{657078}
\definecolor{teaserpanel}{HTML}{F5F5F5}
\usepackage[breaklinks,colorlinks,allcolors=pkured,hyperfootnotes=false,
  bookmarksnumbered,pdfencoding=auto]{hyperref}
\usepackage[capitalize]{cleveref}

\setlist[itemize]{leftmargin=*,topsep=3pt,itemsep=2pt}
\titleformat{\section}{\sffamily\bfseries\large\color{pkured}}{\thesection}{0.9em}{}
\titleformat{\subsection}{\sffamily\bfseries\normalsize\color{pkured}}{\thesubsection}{0.8em}{}
\titlespacing*{\section}{0pt}{14pt plus 3pt minus 2pt}{7pt}
\titlespacing*{\subsection}{0pt}{10pt plus 2pt minus 1pt}{5pt}
\renewcommand{\paragraph}[1]{\par\addvspace{5pt}\noindent\textbf{#1}\enspace}

\renewcommand{\headrulewidth}{0.4pt}
\renewcommand{\headrule}{\hbox to\headwidth{\color{pkured}\leaders\hrule height \headrulewidth\hfill}}
\fancypagestyle{plain}{\fancyhf{}\fancyfoot[C]{\thepage}\renewcommand{\headrulewidth}{0pt}}

\renewenvironment{abstract}{\par\normalfont\normalsize\ignorespaces}{\par}

\makeatletter
\DeclareRobustCommand\onedot{\futurelet\@let@token\@onedot}
\def\@onedot{\ifx\@let@token.\else.\null\fi\xspace}
 
 \def\vs{\emph{vs}\onedot}

\makeatother

%% file: cover.tex
\thispagestyle{empty}
\begingroup
\setlength{\parskip}{2pt}
\vspace*{-8pt}
\noindent
\begin{minipage}[b]{0.68\textwidth}
  \includegraphics[height=24pt]{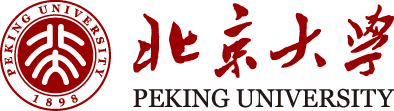}\hspace{18pt}%
  \includegraphics[height=24pt]{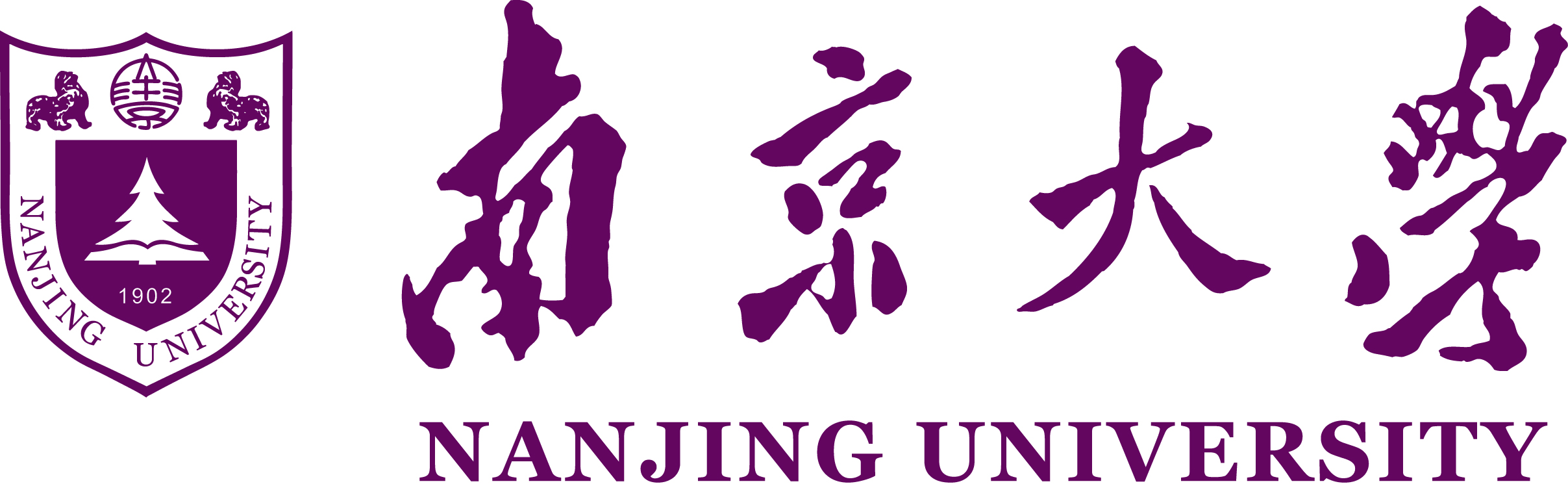}
\end{minipage}%
\begin{minipage}[b]{0.32\textwidth}
  \raggedleft\sffamily\bfseries\small\color{black!70}\reportdate\par\vspace{2pt}
\end{minipage}
\par\vspace{3pt}
{\color{pkured}\hrule height 0.4pt}
\vspace{7pt}
\begin{tcolorbox}[enhanced,colback=pkupale,colframe=pkupale,
  boxrule=0pt,arc=10pt,outer arc=10pt,boxsep=0pt,
  left=16pt,right=16pt,top=9pt,bottom=9pt]
{\sffamily\bfseries\fontsize{16}{18}\selectfont\raggedright\color{pkubrand}
\noindent\resizebox{\linewidth}{!}{Spark-H3: Adaptive Block Sparse Attention for Video Diffusion Models}\par}
\vspace{7pt}
{\centering\sffamily\bfseries\fontsize{10}{12}\selectfont
Zecheng Tang$^{1,*}$, Jiaye Fu$^{1,*}$, Youtian Lin$^{2}$, Xinyao Rao$^{2}$,\\
Liuxuan Sun$^{2}$, Jian Zhang$^{1}$, Yao Yao$^{2}$, Wei Li$^{2,\dagger}$\par}
\vspace{3pt}
{\centering\fontsize{10}{13}\selectfont\raisebox{-2.15pt}{\includegraphics{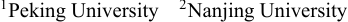}}\par}
\vspace{5pt}
\input{sec/spark_0_abstract}
\vspace{5pt}
{\small $^{*}$Equal contribution.\quad $^{\dagger}$Corresponding author.\par}
\vspace{5pt}
{\small
\noindent\begin{tabular}{@{}c@{\hspace{0.65em}}>{\sffamily\bfseries}l@{\hspace{0.8em}}l@{}}
\faEnvelope & Email: & \href{mailto:lexli@nju.edu.cn}{\nolinkurl{lexli@nju.edu.cn}}\\
\faGithub & Source Code: & \url{https://github.com/zechengtang/Spark-H3}
\end{tabular}\par}
\vspace{3pt}
\noindent\makebox[\linewidth][c]{\includegraphics[width=0.88\linewidth]{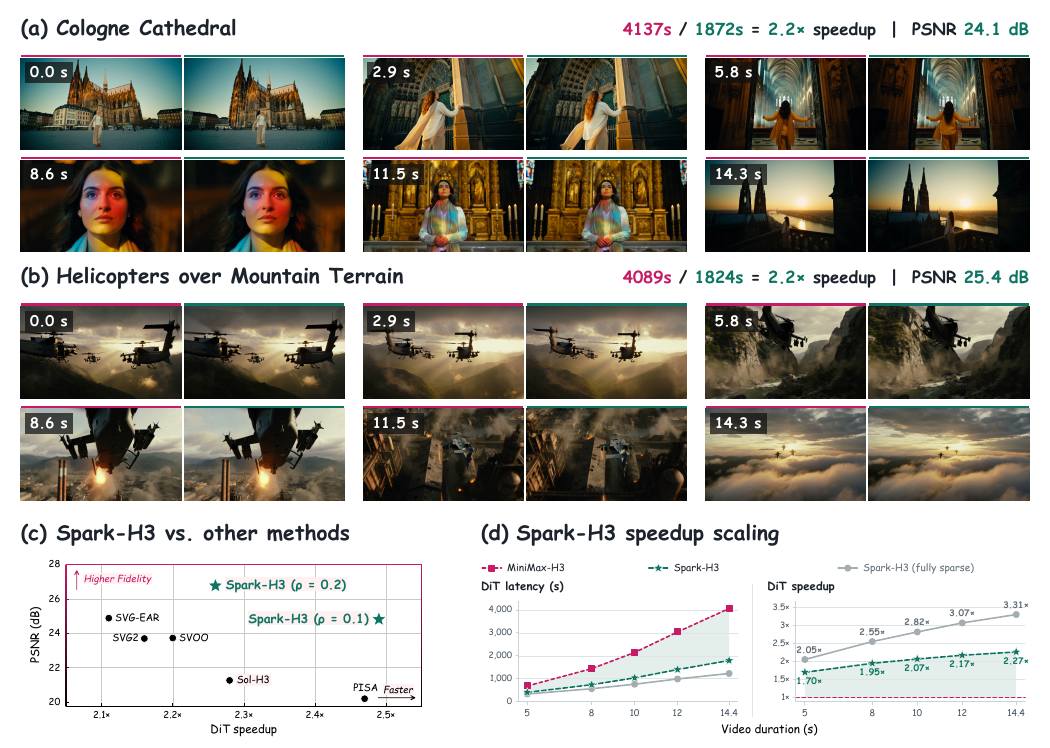}}\par

\begingroup
\captionsetup{hypcap=false,font=footnotesize,skip=3pt}
  \captionof{figure}{\textbf{\textcolor{teaserdense}{MiniMax-H3} and
    \textcolor{teaserspark}{Spark-H3}.}
    (a,b) Qualitative comparisons on 14.4s, $1344\times768$
    videos generated using 20 denoising steps.
    Spark-H3 uses exact-block density $\rho=0.2$ in these examples. (c) DiT speedup \vs PSNR for five state-of-the-art sparse-attention methods and two Spark-H3 configurations ($\rho=0.1,0.2$) under the same inference setting. (d) DiT latency and speedup across video durations. The \textcolor{teaserfullysparse}{fully sparse} Spark-H3 variant omits the 4-step dense warm-up and provides an optimistic upper bound on speedup.
}
  \label{fig:teaser}

\endgroup
\end{tcolorbox}
\endgroup

%% file: sec/spark_0_abstract.tex
\begin{abstract}
Long video sequences make attention a substantial cost in video diffusion models. Block sparse attention (BSA) reduces this cost by computing selected blocks exactly while compressing the remaining blocks. However, fixed token grouping can place tokens with dissimilar attention preferences in the same group, while uniform key averaging can underestimate the attention mass of compressed blocks. We introduce Spark-Attn, a training-free BSA method comprising Spark-Reblock and Spark-Reweight. Spark-Reblock recursively groups tokens by attention preference while preserving regular block shapes for efficient GPU execution.
Spark-Reweight uses the global mean query as an anchor to construct attention-weighted key--value representations and logit biases for the compressed blocks, calibrating their attention mass.
We instantiate Spark-Attn in MiniMax-H3 as Spark-H3. At 10\% exact-block density, Spark-H3 delivers a 2.49$\times$ DiT speedup and reaches 24.81 dB PSNR against the corresponding dense outputs on MiniMax-H3 (20 denoising steps, 345 frames, 1344$\times$768).
Increasing the density to 20\% further improves PSNR to 26.76 dB while still achieving a 2.26$\times$ DiT speedup.
These results demonstrate that Spark-H3 achieves a state-of-the-art quality--efficiency trade-off for training-free BSA on video diffusion models.
\end{abstract}

%% file: sec/spark_1_intro.tex
\section{Introduction}
\label{sec:intro}

Video diffusion transformers (DiTs)~\cite{peebles2023dit} generate videos by repeatedly processing spatiotemporal token sequences throughout the denoising process. Dense self-attention~\cite{vaswani2017attention} compares every query token with every key token, incurring quadratic complexity with respect to sequence length. The FlashAttention family~\cite{dao2022flashattention,dao2024flashattention2,shah2024flashattention3,zadouri2026flashattention4} improves memory efficiency and hardware utilization but retains this dense all-pairs computation. Attention cost therefore grows rapidly with video resolution and duration and can become a major bottleneck at inference time. Reducing this cost while preserving generation quality is important for efficient high-quality video generation.

Recent open-source video diffusion models, including HunyuanVideo~\cite{hunyuanvideo}, the Wan family~\cite{wan}, the LTX-2 family~\cite{ltx2}, and MiniMax-H3~\cite{minimaxh3}, demonstrate the rapid progress of video generation. Within this landscape, we study attention acceleration mainly on MiniMax-H3, whose strong generative capability makes it a practically relevant testbed for assessing whether attention acceleration can provide substantial speedups while preserving output fidelity. Unlike trainable sparse or hybrid attention methods such as VSA and Video DeltaNet~\cite{vsa,vdn}, we consider a training-free setting in which the pretrained parameters and denoising schedule remain unchanged while dense attention is replaced by sparse attention during inference. Because the model is trained with dense attention, preserving its inference behavior helps limit the training--inference discrepancy introduced by this replacement. Our goal is therefore to reduce attention latency while keeping the generated outputs close to their dense-attention counterparts.

Early training-free sparse attention methods for video diffusion models exploit structured spatial--temporal redundancy through head-specific patterns or localized sliding tiles~\cite{svg,sta}. Prior block sparse attention (BSA) methods~\cite{spargeattn,xattention,pisa,solattn} often use fixed token grouping, which can place tokens with dissimilar attention preferences in the same group. Cluster sparse attention (CSA) methods~\cite{svg2,svoo,svgear} instead regroup tokens adaptively to better align grouping with attention preferences. Within BSA, fixed grouping spreads important query--key interactions across more blocks, making them harder to capture at a fixed exact-block density. Meanwhile, uniform key averaging replaces the mean of the exponentiated logits within a block with the exponential of their mean. The resulting Jensen gap can cause compressed blocks to receive too little attention mass. We identify these two failure modes as \emph{preference mismatch} and \emph{compression bias}, respectively.

\begin{figure}[t]
  \centering
  \includegraphics[width=0.78\textwidth]{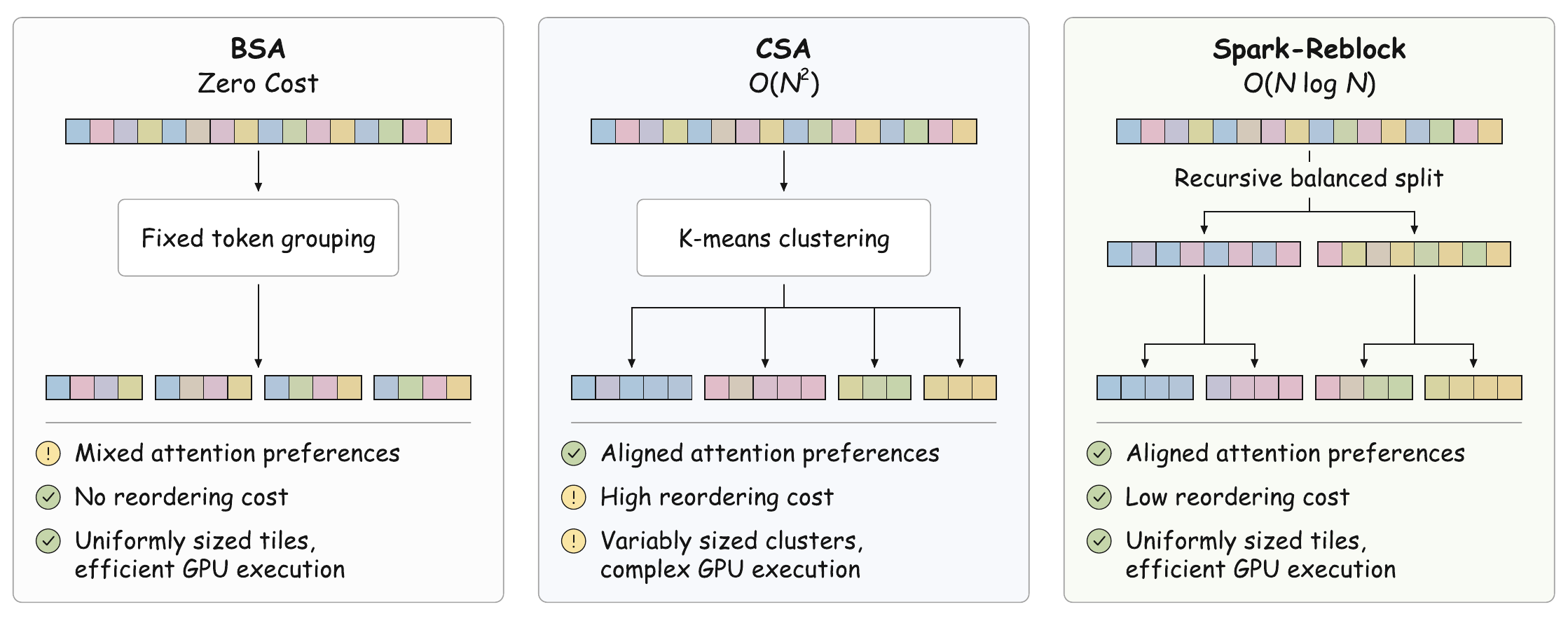}
  \caption{\textbf{BSA, CSA, and Spark-Reblock.}
  Colors denote attention preferences, and $N$ denotes the number of video tokens.
  Spark-Reblock uses recursive balanced splits to construct preference-aligned,
  uniformly sized tiles in $O(N\log N)$.}
  \label{fig:reblock-tiling}
\end{figure}

We introduce Spark-Attn, a training-free BSA method that addresses both failure modes. As illustrated in Fig.~\ref{fig:reblock-tiling}, Spark-Reblock uses recursive balanced splits to group query and key--value (KV) tokens with similar attention preferences while preserving regular block shapes. Spark-Reweight uses the global mean query as an anchor to construct attention-weighted KV representations and logit biases for the compressed blocks, calibrating their attention mass. We instantiate Spark-Attn in MiniMax-H3 as Spark-H3 without parameter training, adapters, or learned auxiliary modules. Our key contributions are as follows:

\begin{itemize}
  \setlength{\itemsep}{1pt}
  \setlength{\parskip}{0pt}
  \setlength{\parsep}{0pt}
  \setlength{\topsep}{2pt}
  \item We introduce Spark-Attn, a training-free BSA for video diffusion models, which comprising Spark-Reblock and Spark-Reweight.
  \item We propose Spark-Reblock, a preference-aware hierarchical partitioning method that aligns token grouping with attention structure while preserving regular block shapes for efficient GPU execution.
  \item We develop Spark-Reweight, which uses a global query anchor to calibrate the attention mass of compressed blocks.
  \item We instantiate Spark-Attn in MiniMax-H3 as Spark-H3 and evaluate it across several video diffusion models, demonstrating consistent state-of-the-art quality--efficiency trade-offs for training-free BSA on video diffusion models.
\end{itemize}

%% file: sec/spark_2_related.tex
\section{Related work}
\label{sec:related}

\paragraph{Training-free sparse attention.}
Training-free sparse attention accelerates pretrained video diffusion models without updating model parameters or training auxiliary modules. Early pattern-based methods exploit spatial--temporal redundancy using fixed layouts or by selecting head-specific layouts from predefined pattern families. Sparse VideoGen (SVG)~\cite{svg} profiles spatial and temporal attention heads online, Sliding Tile Attention (STA)~\cite{sta} computes dense attention within local spatiotemporal windows, and PAROAttention~\cite{paroattention} selects per-head layouts from six spatiotemporal axis permutations through offline profiling.
Block sparse attention (BSA) makes regular blocks the unit of routing, computing selected blocks exactly and handling unselected blocks through a compressed branch, with dropping as the zero-contribution special case. SpargeAttention~\cite{spargeattn} and XAttention~\cite{xattention} focus on block-importance estimation, while AdaSpa~\cite{adaspa} searches block masks at selected denoising steps and reuses sparse patterns across steps. PISA~\cite{pisa} and Sol-Attn~\cite{solattn} approximate the contributions of unselected blocks through block-wise Taylor correction and compressed KV representations, respectively.
Cluster sparse attention (CSA) methods instead use adaptive token grouping to expose sparse structure. SVG2~\cite{svg2} uses semantic-aware permutation, SVOO~\cite{svoo} combines offline profiling with online query--key co-clustering, and SVG-EAR~\cite{svgear} adds error-aware routing and centroid compensation.
Spark-Reblock combines adaptive token grouping with regular block shapes, while Spark-Reweight uses the global mean query as an anchor to calibrate the attention mass of compressed blocks.

\paragraph{Training-based attention adaptation.}
Training-based approaches either train models natively with efficient attention or adapt pretrained models. Representative native architectures include MoBA~\cite{moba}, which formulates block routing as a mixture-of-experts process, and NSA~\cite{nsa}, which combines coarse-grained token compression with fine-grained token selection. VMoBA~\cite{vmoba} and Log-linear Sparse Attention~\cite{llsa} extend trainable sparse attention to video DiTs, while SANA-Video~\cite{sanavideo} adopts a linear DiT with autoregressive state for long-video generation. VSA~\cite{vsa} supports both pretraining with sparse attention and adapting pretrained models.
Post-training adaptation methods include Veda~\cite{veda} and SpargeAttention2~\cite{spargeattn2}, which adapt BSA, and SLA and SLA2~\cite{sla,sla2}, which combine sparse and linear attention. VORTA~\cite{vorta} learns a lightweight timestep-conditioned router over three predefined attention patterns while freezing the pretrained backbone. For MiniMax-H3 specifically, FastH3~\cite{fasth3,fasth3v2} combines trained sparse attention with few-step distillation, whereas Video DeltaNet~\cite{vdn} learns a hybrid of local softmax attention and linear memory. Because these approaches train attention parameters or auxiliary modules for a target model, applying them to another LoRA of the same model or to a different model may require further optimization. Spark-Attn instead restructures attention computation itself, requiring neither model-specific adaptation nor retraining when applied to different LoRAs of the same model or to different models.

%% file: sec/spark_3_method.tex
\section{Method}
\label{sec:method}

\begin{figure}[t]
\centering
\includegraphics[width=\textwidth]{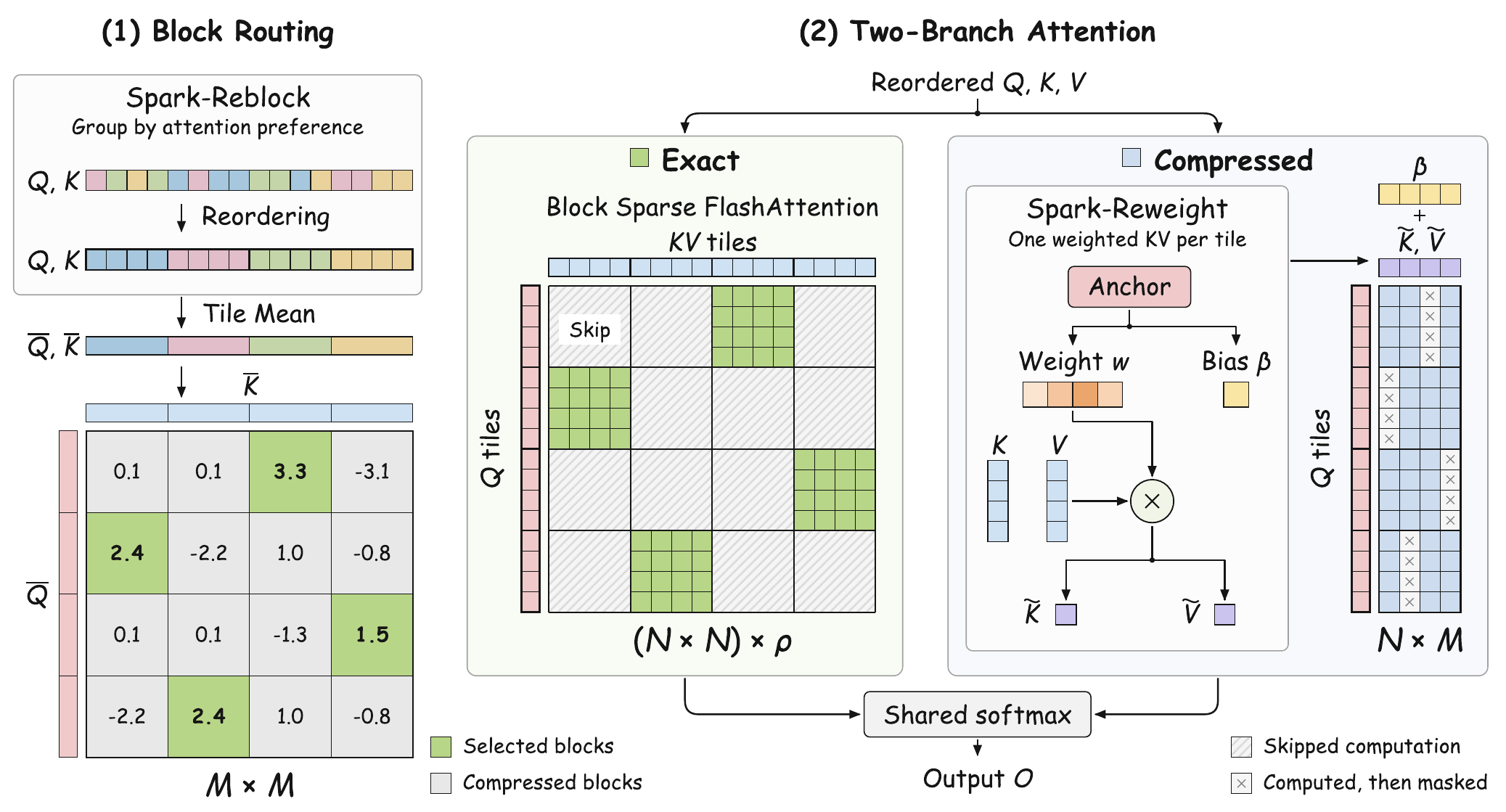}
\caption{\textbf{Spark-Attn overview.}
\textbf{(1) Block routing:}
Spark-Reblock groups query and key--value (KV) tokens into
preference-aware, uniformly sized tiles; tile means guide selection of
a fraction $\rho$ of KV tiles per query tile.
\textbf{(2) Two-branch attention:}
Selected blocks are computed exactly, while the rest are approximated
using Spark-Reweight's weighted KV representatives and logit biases.
Both branches share one softmax normalization, with compressed entries
for selected blocks masked to avoid double counting.
$N$, $M$, and $\rho$ denote the number of video tokens, the number of
KV tiles, and the exact-block density, respectively.}
\label{fig:pipeline}
\end{figure}

Spark-Attn accelerates dense self-attention without adapting the pretrained parameters. We call a group of query or KV tokens a \emph{tile}, and the query--key score matrix between a query tile and a KV tile an \emph{attention block}. Given the query, key, and value matrices $Q,K,V$ for one attention head, Spark-Attn uses a block router to select a configured fraction $\rho$, for example, 20\%, of candidate KV tiles per query tile for exact computation; a compressed branch approximates the remaining blocks. As shown in Fig.~\ref{fig:pipeline}, Spark-Reblock addresses preference mismatch by constructing uniformly sized tiles with similar attention preferences, allowing the block router to capture more attention mass at the same exact-block density $\rho$. Spark-Reweight addresses compression bias by constructing attention-weighted KV representatives and a logit bias for each unselected KV tile. The two branches share one softmax normalization, and the output is restored to the original query order. We next formalize the two-branch computation, then detail preference-aware tile construction in Spark-Reblock and anchor-based compression in Spark-Reweight.

\subsection{Two-Branch Block Sparse Attention}
\label{sec:two-branch-attention}

For one attention head over $N$ video tokens, let $Q,K,V\in\mathbb{R}^{N\times d}$ be the query, key, and value matrices, with head dimension $d$. For a query $q_i$ and a key $k_j$, define the scaled dot-product score as $s(q_i,k_j)=q_i^\top k_j/\sqrt d$. Dense attention~\cite{vaswani2017attention} exponentiates these scores to weight the corresponding values $v_j$. For subsequent block-wise accumulation, we equivalently write dense attention as
\begin{equation}
\bigl(n(q_i),z(q_i)\bigr)=\sum_{j=1}^{N}e^{s(q_i,k_j)}(v_j,1),
\; o(q_i)=\frac{n(q_i)}{z(q_i)}.
\label{eq:dense-attention}
\end{equation}
Here $n(q_i)$ is the value-weighted numerator and $z(q_i)$ is the total attention mass.

\paragraph{Tiles and attention blocks.}
The scores form a matrix $S=QK^\top/\sqrt d\in\mathbb{R}^{N\times N}$. FlashAttention~\cite{dao2022flashattention} computes these scores block by block on chip while updating the numerator and denominator in Eq.~\eqref{eq:dense-attention}. Let $\mathcal T_I^Q$ and $\mathcal T_J^K$ denote query and key tiles of sizes $b_q$ and $b_{kv}$, respectively. The values aligned with the latter form $\mathcal T_J^V$, giving the KV tile $\mathcal T_J^{KV}=(\mathcal T_J^K,\mathcal T_J^V)$. Their scaled matrix product $\mathcal T_I^Q(\mathcal T_J^K)^\top/\sqrt d\in\mathbb{R}^{b_q\times b_{kv}}$ defines an attention block, whose entries are the individual query--key scores.

\paragraph{Block routing.}
Block sparse attention (BSA) reduces the cost of Eq.~\eqref{eq:dense-attention} by computing token-level scores only in selected blocks. Following MoBA~\cite{moba}, we use tile means for block routing. For query tile $I$ and KV tile $J$, let $\bar q_I$ and $\bar k_J$ denote their respective mean query and key; the router scores the pair by $r_{IJ}=s(\bar q_I,\bar k_J)$. For each query tile $I$, it selects a set $\mathcal S_I$ containing the highest-scoring fraction $\rho$ of the $M$ candidate KV tiles, which we call the configured exact-block density. Every query in tile $I$ shares this selection. The tile layout therefore determines which token-level scores can be computed together under a given exact-block budget, directly affecting how much dense attention mass the selected blocks capture.

\paragraph{Approximating unselected blocks.}
Unselected blocks can still carry substantial attention mass in aggregate. We therefore use a two-branch formulation in which selected blocks contribute their token-level scores exactly, whereas unselected blocks contribute through a compressed approximation.

For a query $q_i\in\mathcal T_I^Q$, the exact contribution of a KV tile $J$ to Eq.~\eqref{eq:dense-attention} is $(n_J(q_i),m_J(q_i))=\sum_{j\in J}e^{s(q_i,k_j)}(v_j,1)$, where $j\in J$ indexes the token positions in the tile. The exact branch computes this contribution from token-level scores for each selected tile. For each unselected tile, the compressed branch approximates its contribution by $(\widehat n_J(q_i),\widehat m_J(q_i))$. Accumulating the two branches gives
\begin{equation}
\begin{aligned}
\bigl(\widehat n(q_i),\widehat z(q_i)\bigr)
&=\sum_{J\in\mathcal S_I}\sum_{j\in J}e^{s(q_i,k_j)}(v_j,1)\\
&\quad+\sum_{J\notin\mathcal S_I}
\bigl(\widehat n_J(q_i),\widehat m_J(q_i)\bigr).
\end{aligned}
\label{eq:two-branch-accumulation}
\end{equation}

Each block contributes through exactly one branch, and compressed entries corresponding to selected blocks are masked to avoid double counting. The final output is $\widehat o(q_i)=\widehat n(q_i)/\widehat z(q_i)$, following Eq.~\eqref{eq:dense-attention}. Because both branches contribute to $\widehat z(q_i)$, compressed blocks also affect the normalized weights of the selected keys.

This two-branch formulation leaves two design choices: the tile layout used for block routing and the approximation of unselected blocks. We address them next with Spark-Reblock and Spark-Reweight, respectively.

\subsection{Spark-Reblock}

Fixed token grouping couples block routing to the spatiotemporal layout rather than to the attention structure. When a tile mixes tokens with dissimilar attention preferences, the tile means used by the block router become less representative of the token-level scores, increasing the chance that high-mass blocks are unselected at a given exact-block density. Spark-Reblock addresses this preference mismatch by using the current $Q$ and $K$ to construct a token permutation and uniformly sized tiles that group tokens with similar attention preferences. It thus combines the adaptive grouping of cluster sparse attention (CSA) with the regular block shapes of block sparse attention (BSA). As illustrated in Fig.~\ref{fig:reblock-tiling}, its construction is analogous to median-pivot quicksort: each node uses a landmark-derived direction to define scalar split scores, and equal-capacity splits partition them at the median rank. The objective, however, is uniformly sized tiles rather than a globally sorted sequence.

\paragraph{Attention preference.}
We define \emph{attention preference} as the ranking induced by scaled dot-product scores: all keys for a query, or all queries for a key. For a query $q$, the corresponding row of $S$ lists its scores against all keys. Normalizing this row gives $u_Q(q)=Kq/\lVert Kq\rVert_2$. Rather than comparing raw query features by Euclidean distance, we use cosine similarity between normalized score vectors as a surrogate for attention-preference similarity. The non-centered second moment $M_K=K^\top K/N$ expresses this comparison as a cosine distance under a key-induced Mahalanobis-type inner product:
\begin{equation}
\begin{aligned}
d_Q(q,q')&=1-u_Q(q)^\top u_Q(q')\\
&=1-\cos(M_K^{1/2}q,M_K^{1/2}q').
\end{aligned}
\label{eq:metric}
\end{equation}
The distance $d_K$ between keys follows symmetrically by comparing their interactions with the current query set using $M_Q=Q^\top Q/N$. Before recursion, we estimate $\widehat M$ from sampled tokens and reuse the resulting metric throughout the partition tree, avoiding construction of the full attention matrix.

\begin{algorithm}[!t]
\caption{Recursive tile construction in Spark-Reblock}
\label{alg:spark-reblock}
\small
\begin{algorithmic}[1]
\State \textbf{Inputs:} token features $\{x_i\}_{i\in G_0}$, tile size $b_\star$,
estimated second moment $\widehat M$, and child capacities in multiples of $b_\star$.
\State \textbf{Output:} ordered list of tiles.
\State \textbf{Init:} $z_i\gets\widehat M^{1/2}x_i$ for each $i\in G_0$
\State \Return \Call{Split}{$G_0$}
\Procedure{Split}{$G$}
    \If{$|G|=b_\star$}
        \State \Return $[G]$
    \EndIf
    \State Read the prescribed left child capacity $c_L$ for $G$
    \State Initialize $\mu_{G,-},\mu_{G,+}$ from a dissimilar landmark pair
    \State Refine both representatives on the local landmarks
    \State Compute $\Delta_G(x_i)$ for $i\in G$ by Eq.~\eqref{eq:split-margin}
    \State $G_L\gets$ indices of the $c_L$ smallest $(\Delta_G(x_i),i)$
    \State $G_R\gets G\setminus G_L$
    \State \Return \Call{Split}{$G_L$} followed by \Call{Split}{$G_R$}
\EndProcedure
\end{algorithmic}
\end{algorithm}

\paragraph{Node-specific split direction.}
A single global split direction may not capture the preference structure remaining within every node. Each internal node $G$ therefore fits two representatives $\mu_{G,-}$ and $\mu_{G,+}$ from its local landmarks. For a token $x$ with transformed feature $z_x=\widehat M^{1/2}x$, we define the scalar split score
\begin{equation}
\Delta_G(x)=\cos(z_x,\mu_{G,+})-\cos(z_x,\mu_{G,-}).
\label{eq:split-margin}
\end{equation}
The difference between the two unit-normalized representatives gives the node's split direction, and the resulting scores provide the ordering used for rank partitioning.

\paragraph{Recursive rank partitioning.}
Recursive splitting terminates when $|G|=b_\star$, yielding uniformly sized tiles. Child capacities are imposed through rank cutoffs, reducing to median-rank splits when the two capacities are equal. Algorithm~\ref{alg:spark-reblock} summarizes the procedure.

\paragraph{Batched construction.}
A full partitioning round batches four binary levels: a landmark tree constructs 15 split directions for 16 child groups, and one batched matrix product evaluates their split scores before rank cutoffs and token permutation. With fixed feature dimension and landmark budget, linear-time rank selection gives $O(N\log N)$ construction time in the token count. Spark-H3 uses 32 landmarks, two refinement updates per split, and 64 tokens per tile, by default.

\definecolor{rankfirst}{RGB}{165,211,143}
\definecolor{ranksecond}{RGB}{198,228,183}
\definecolor{rankthird}{RGB}{228,241,220}

\paragraph{Partitioning and routing.}
Spark-H3 derives a single partition from the keys using $d_K$ and applies its permutation to queries, keys, and values, avoiding a second partitioning pass. The attention output is restored to the original query order. Spark-H3$^{\dagger}$ instead uses separate query and key--value partitions constructed with $d_Q$ and $d_K$, respectively. After reblocking, the block router scores each query--KV tile pair using the query and key tile means and selects the top fraction $\rho$ of KV tiles per query tile for the exact branch.

\subsection{Spark-Reweight}
Consider a KV tile $J$ left unselected for query tile $\mathcal T_I^Q$, i.e., $J\notin\mathcal S_I$. For any query $q\in\mathcal T_I^Q$, the compressed branch must approximate both the tile's total attention mass $m_J(q)$ and its value-weighted numerator $n_J(q)$. Mean-pooled KV compression uses $\bar k_J=b_{kv}^{-1}(\mathcal T_J^K)^\top\mathbf 1$ and estimates the log-mass $\ell_J(q)=\log m_J(q)$ by $s(q,\bar k_J)+\log b_{kv}$. Jensen's inequality gives the nonnegative gap
\begin{equation}
\ell_J(q)-\bigl(\log b_{kv}+s(q,\bar k_J)\bigr)\ge 0,
\label{eq:jensen-gap}
\end{equation}
which can make compressed blocks receive too little attention mass. Spark-Reweight addresses this compression bias by constructing attention-weighted key and value representatives together with a scalar logit bias, using the global mean query as a shared anchor.

Spark-Reweight first computes the global mean query $\bar q=N^{-1}\sum_{i=1}^N q_i$ for each attention head. For any query $q$, let $\mathbf e_J(q)=\exp(\mathcal T_J^Kq/\sqrt d)\in\mathbb R^{b_{kv}}$ denote the exponentiated scores within KV tile $J$, with the exponential applied elementwise. Evaluating these scores at the anchor and normalizing within the tile gives the pooling weights
\begin{equation}
\mathbf w_J(\bar q)=\frac{\mathbf e_J(\bar q)}{m_J(\bar q)},
\qquad
m_J(\bar q)=\mathbf 1^\top\mathbf e_J(\bar q).
\label{eq:weights}
\end{equation}
These weights are shared by all queries and assign greater weight to keys with higher scores at the anchor.
Using these weights, Spark-Reweight constructs the key and value representatives $\widetilde k_J=(\mathcal T_J^K)^\top\mathbf w_J(\bar q)$ and $\widetilde v_J=(\mathcal T_J^V)^\top\mathbf w_J(\bar q)$. It pairs them with the logit bias
\begin{equation}
\beta_J=\log m_J(\bar q)-s(\bar q,\widetilde k_J).
\label{eq:bias}
\end{equation}
The bias measures the difference between the tile's exact log-mass and the score of its weighted key representative at the anchor.
For an unselected block, the resulting representation $(\widetilde k_J,\widetilde v_J,\beta_J)$ gives
\begin{equation}
\widehat m_J(q)=e^{s(q,\widetilde k_J)+\beta_J},
\qquad
\widehat n_J(q)=\widehat m_J(q)\widetilde v_J.
\label{eq:reweight}
\end{equation}
These contributions enter Eq.~\eqref{eq:two-branch-accumulation} directly. The representatives and bias are computed once from the current $Q,K,V$ and reused across queries, while the estimated mass remains query-dependent through $s(q,\widetilde k_J)$.

The weight--bias construction admits a simple mathematical interpretation as a first-order approximation to the tile's log-sum-exp. Mean-pooled compression corresponds to expanding $\ell_J$ at a zero anchor: it uses uniform weights and $\beta_J=\log b_{kv}$, giving $T_0\ell_J(q)=s(q,\bar k_J)+\log b_{kv}$ and hence the Jensen gap in Eq.~\eqref{eq:jensen-gap}. Spark-Reweight instead expands $\ell_J$ at the observed mean query. Since $\nabla\ell_J(\bar q)=\widetilde k_J/\sqrt d$, the first-order expansion is
\begin{equation}
\begin{aligned}
T_{\bar q}\ell_J(q)&=\ell_J(\bar q)+(q-\bar q)^\top\nabla\ell_J(\bar q)\\
&=s(q,\widetilde k_J)+\beta_J.
\end{aligned}
\label{eq:anchor-taylor}
\end{equation}
Equation~\eqref{eq:reweight} therefore gives exactly $\widehat m_J(q)=e^{T_{\bar q}\ell_J(q)}$. At $\bar q$, the tangent matches both the log-mass and its gradient, and $\widetilde v_J$ equals the tile's normalized value aggregate; consequently, $\widehat m_J(\bar q)=m_J(\bar q)$ and $\widehat n_J(\bar q)=n_J(\bar q)$. Away from the anchor, the construction remains a first-order approximation rather than an exactness guarantee.

%% file: sec/spark_4_experiments.tex
\section{Experiments}
\label{sec:experiments}

\subsection{Evaluation protocol}

\paragraph{Benchmark.}
We evaluate all sparse-attention methods on the VBench benchmark suite~\cite{vbench}. Following prior work~\cite{svg2,svgear,svoo}, we use GPT-enhanced prompts from the Penguin Benchmark. For MiniMax-H3, we adapt these prompts to its prompt template~\cite{minimaxh3}.

\paragraph{Metrics.}
We evaluate quality and efficiency. For efficiency, we measure DiT latency, excluding text encoding and VAE decoding, and report speedup over the DiT latency of dense-attention results under identical generation settings. For quality, we use PSNR, SSIM~\cite{wang2004ssim}, and LPIPS~\cite{lpips} to measure fidelity to dense-attention outputs, capturing the inference discrepancy introduced by sparse-attention replacement. We also report the average VBench score over the five core dimensions used in prior work~\cite{svg2,svgear,svoo,pisa}.

\paragraph{Baselines.}
We denote the original model outputs as Dense and compare Spark-H3 and Spark-H3$^{\dagger}$ with five recent state-of-the-art training-free sparse-attention baselines: SVG2~\cite{svg2}, SVG-EAR~\cite{svgear}, SVOO~\cite{svoo}, PISA~\cite{pisa}, and Sol-Attn~\cite{solattn}.

\begin{table}[!t]
\centering\small
\setlength{\tabcolsep}{3pt}
\caption{\textbf{Quantitative Results on MiniMax-H3.} Green shading marks the \colorbox{rankfirst}{first}, \colorbox{ranksecond}{second}, and \colorbox{rankthird}{third} distinct scores among methods. $^{\dagger}$ uses separate query and KV partitions.}
\label{tab:main}
\begin{tabular*}{\textwidth}{@{\extracolsep{\fill}}llccccccc}
\toprule
\textbf{Method} & \textbf{Category} & \textbf{Density} & \textbf{DiT latency (s) $\downarrow$} & \textbf{Speedup $\uparrow$} & \textbf{PSNR (dB) $\uparrow$} & \textbf{SSIM $\uparrow$} & \textbf{LPIPS $\downarrow$} & \textbf{VBench $\uparrow$} \\
\midrule
Dense & -- & 100\% & 4061.8 & 1.00$\times$ & $\infty$ & 1.00 & 0.00 & 84.38 \\
\midrule
SVG2~\cite{svg2} & CSA & $\approx17.8\%$ & 1876.1 & 2.16$\times$ & 23.70 & 0.79 & 0.15 & 84.39 \\
SVG-EAR~\cite{svgear} & CSA & $\approx17.8\%$ & 1929.3 & 2.11$\times$ & \cellcolor{rankthird}24.88 & 0.82 & 0.13 & \cellcolor{rankthird}84.50 \\
SVOO~\cite{svoo} & CSA & $\approx19.7\%$ & 1847.6 & 2.20$\times$ & 23.73 & 0.79 & 0.14 & \cellcolor{ranksecond}84.56 \\
\midrule
PISA~\cite{pisa} & BSA & 10\% & \cellcolor{ranksecond}1643.4 & \cellcolor{ranksecond}2.47$\times$ & 20.20 & 0.71 & 0.23 & \cellcolor{rankfirst}\textbf{84.60} \\
Sol-H3~\cite{solattn} & BSA & $\approx20.8\%$ & \cellcolor{rankthird}1781.9 & \cellcolor{rankthird}2.28$\times$ & 21.26 & 0.74 & 0.19 & 84.47 \\
\addlinespace[3pt]
Spark-H3 & BSA & 10\% & \cellcolor{rankfirst}\textbf{1629.6} & \cellcolor{rankfirst}\textbf{2.49}$\times$ & 24.81 & \cellcolor{rankthird}0.83 & \cellcolor{rankthird}0.12 & \cellcolor{ranksecond}84.56 \\
Spark-H3 & BSA & 20\% & 1796.1 & 2.26$\times$ & \cellcolor{ranksecond}26.76 & \cellcolor{ranksecond}0.86 & \cellcolor{ranksecond}0.09 & \cellcolor{rankthird}84.50 \\
Spark-H3$^{\dagger}$ & BSA & 20\% & 1818.0 & 2.23$\times$ & \cellcolor{rankfirst}\textbf{26.89} & \cellcolor{rankfirst}\textbf{0.87} & \cellcolor{rankfirst}\textbf{0.08} & \cellcolor{ranksecond}84.56 \\
\bottomrule
\end{tabular*}
\vspace{8pt}

\caption{\textbf{Quantitative Results on LTX-2.5 (High-Resolution Stage).} Notation and shading follow Table~\ref{tab:main}.}
\label{tab:ltx25}
\begin{tabular*}{\textwidth}{@{\extracolsep{\fill}}llccccccc}
\toprule
\textbf{Method} & \textbf{Category} & \textbf{Density} & \textbf{DiT latency (s) $\downarrow$} & \textbf{Speedup $\uparrow$} & \textbf{PSNR (dB) $\uparrow$} & \textbf{SSIM $\uparrow$} & \textbf{LPIPS $\downarrow$} & \textbf{VBench $\uparrow$} \\
\midrule
Dense & -- & 100\% & 1223.6 & 1.00$\times$ & $\infty$ & 1.00 & 0.00 & 82.10 \\
\midrule
SVG2 & CSA & $\approx54.0\%$ & 869.6 & 1.41$\times$ & 30.80 & \cellcolor{rankthird}0.94 & 0.09 & 82.30 \\
SVG-EAR & CSA & $\approx53.9\%$ & 878.3 & 1.39$\times$ & \cellcolor{rankthird}34.20 & \cellcolor{ranksecond}0.96 & \cellcolor{rankthird}0.06 & 82.38 \\
SVOO & CSA & $\approx56.0\%$ & 891.9 & 1.37$\times$ & 29.90 & 0.93 & 0.10 & \cellcolor{rankfirst}\textbf{82.93} \\
\midrule
PISA & BSA & 10\% & \cellcolor{ranksecond}313.4 & \cellcolor{ranksecond}3.90$\times$ & 29.20 & 0.92 & 0.12 & 82.00 \\
Sol-Attn & BSA & $\approx18.5\%$ & \cellcolor{rankthird}385.7 & \cellcolor{rankthird}3.17$\times$ & 24.72 & 0.88 & 0.21 & 81.21 \\
\addlinespace[3pt]
Spark-Attn & BSA & 10\% & \cellcolor{rankfirst}\textbf{312.0} & \cellcolor{rankfirst}\textbf{3.92}$\times$ & 30.84 & 0.93 & 0.09 & 82.38 \\
Spark-Attn & BSA & 20\% & 415.1 & 2.95$\times$ & 32.68 & \cellcolor{rankthird}0.94 & 0.07 & \cellcolor{rankthird}82.39 \\
Spark-Attn & BSA & 50\% & 736.9 & 1.66$\times$ & \cellcolor{rankfirst}\textbf{36.79} & \cellcolor{rankfirst}\textbf{0.97} & \cellcolor{rankfirst}\textbf{0.04} & 82.33 \\
Spark-Attn$^{\dagger}$ & BSA & 50\% & 737.8 & 1.66$\times$ & \cellcolor{ranksecond}35.16 & \cellcolor{ranksecond}0.96 & \cellcolor{ranksecond}0.05 & \cellcolor{ranksecond}82.44 \\
\bottomrule
\end{tabular*}
\end{table}

\subsection{Quantitative Results}

We evaluate Spark-Attn on MiniMax-H3~\cite{minimaxh3} and LTX-2.5~\cite{ltx2}, two leading video diffusion models. These experiments assess quality--efficiency trade-offs across distinct model architectures and generation settings.

\paragraph{Quantitative Results on MiniMax-H3.} Table~\ref{tab:main} summarizes the MiniMax-H3 results. We generate 14.4s videos at $1344\times768$ resolution using a 20-step denoising schedule. All sparse-attention methods use four initial dense denoising steps as warm-up before switching to sparse attention. At 10\% exact-block density, Spark-H3 achieves the lowest DiT latency and a 2.49$\times$ speedup, with a PSNR within 0.07 dB of the best baseline result. Increasing the density to 20\% improves PSNR to 26.76 dB while still achieving a 2.26$\times$ speedup. Spark-H3 improves PSNR over two BSA methods at similar density, with gains of 4.61 dB over PISA at 10\% and 5.50 dB over Sol-H3 at 20\%. At 20\% exact-block density, both Spark-H3 variants achieve higher PSNR and SSIM, lower LPIPS, and lower DiT latency than all three CSA baselines. These results establish a state-of-the-art quality--efficiency trade-off among the evaluated methods.

\paragraph{Quantitative Results on LTX-2.5.} LTX-2.5 uses a two-stage denoising process. We generate 10s videos at $3840\times2176$ resolution using an 11-step schedule: 8 steps at $1920\times1088$, followed by 3 steps at the target resolution after latent upsampling. Following Sol-Attn~\cite{solattn}, we keep the first stage dense and apply sparse-attention methods only to the second stage. Table~\ref{tab:ltx25} therefore reports DiT latency of the second stage, excluding the first stage and latent upsampling.

Spark-Attn achieves a strong quality--efficiency trade-off on LTX-2.5. At 10\% density, it matches PISA's DiT latency while improving PSNR by 1.64 dB, achieving a 3.92$\times$ speedup. Increasing the density to 20\% and 50\% progressively improves PSNR. At 50\% density, Spark-Attn achieves higher PSNR and lower latency than all three CSA baselines. Together with the MiniMax-H3 results, these experimental results support the applicability of Spark-Attn across distinct video diffusion models.

\begin{table}[t]
\centering\small
\setlength{\tabcolsep}{3pt}
\caption{\textbf{Quantitative Results on MiniMax-H3-Turbo.} Notation and shading follow Table~\ref{tab:main}.}
\label{tab:lightx2v}
\begin{tabular*}{\textwidth}{@{\extracolsep{\fill}}llccccccc}
\toprule
\textbf{Method} & \textbf{Category} & \textbf{Density} & \textbf{DiT latency (s) $\downarrow$} & \textbf{Speedup $\uparrow$} & \textbf{PSNR (dB) $\uparrow$} & \textbf{SSIM $\uparrow$} & \textbf{LPIPS $\downarrow$} & \textbf{VBench $\uparrow$} \\
\midrule
Dense & -- & 100\% & 1625.0 & 1.00$\times$ & $\infty$ & 1.00 & 0.00 & 84.32 \\
\midrule
SVG2 & CSA & $\approx17.6\%$ & 846.7 & 1.92$\times$ & 24.01 & 0.78 & 0.15 & 84.32 \\
SVG-EAR & CSA & $\approx17.6\%$ & 864.8 & 1.88$\times$ & \cellcolor{rankthird}25.01 & \cellcolor{ranksecond}0.81 & \cellcolor{ranksecond}0.13 & 84.31 \\
SVOO & CSA & $\approx19.1\%$ & 829.0 & 1.96$\times$ & 24.33 & \cellcolor{rankthird}0.80 & \cellcolor{rankthird}0.14 & \cellcolor{rankfirst}\textbf{84.54} \\
\midrule
PISA & BSA & 10\% & \cellcolor{ranksecond}718.7 & \cellcolor{ranksecond}2.26$\times$ & 21.28 & 0.71 & 0.22 & \cellcolor{rankthird}84.39 \\
Sol-H3 & BSA & $\approx20.9\%$ & \cellcolor{rankthird}775.6 & \cellcolor{rankthird}2.10$\times$ & 22.44 & 0.75 & 0.17 & \cellcolor{ranksecond}84.43 \\
\addlinespace[3pt]
Spark-H3 & BSA & 10\% & \cellcolor{rankfirst}\textbf{717.2} & \cellcolor{rankfirst}\textbf{2.27}$\times$ & 24.80 & \cellcolor{ranksecond}0.81 & \cellcolor{ranksecond}0.13 & 84.37 \\
Spark-H3 & BSA & 20\% & 780.2 & 2.08$\times$ & \cellcolor{ranksecond}26.48 & \cellcolor{rankfirst}\textbf{0.85} & \cellcolor{rankfirst}\textbf{0.09} & 84.36 \\
Spark-H3$^{\dagger}$ & BSA & 20\% & 785.0 & 2.07$\times$ & \cellcolor{rankfirst}\textbf{26.65} & \cellcolor{rankfirst}\textbf{0.85} & \cellcolor{rankfirst}\textbf{0.09} & 84.34 \\
\bottomrule
\end{tabular*}
\end{table}

\begin{table}[!t]
\centering\small
\setlength{\tabcolsep}{1pt}
\caption{\textbf{Ablation of token grouping methods}. Attention-mass recall is the fraction of dense-softmax attention mass captured by selected blocks (mean $\pm$ sample SD).}
\label{tab:attn-recall}
\begin{tabular*}{\columnwidth}{@{\extracolsep{\fill}}@{}>{\centering\arraybackslash}p{0.15\columnwidth}>{\centering\arraybackslash}p{0.16\columnwidth}>{\centering\arraybackslash}p{0.21\columnwidth}>{\centering\arraybackslash}p{0.21\columnwidth}>{\centering\arraybackslash}p{0.21\columnwidth}@{}}
\toprule
\textbf{Spark-Reblock} & \textbf{Density} & \textbf{DiT Latency (s) $\downarrow$} & \textbf{Recall (\%) $\uparrow$} & \textbf{PSNR (dB) $\uparrow$} \\
\midrule
 & 10\% & 1598.6 & $61.36\pm2.71$ & 20.30 \\
\checkmark & 10\% & 1629.6 & $79.69\pm0.91$ & 24.81 \\
\addlinespace[2pt]
 & 20\% & 1763.2 & $74.05\pm2.12$ & 21.65 \\
\checkmark & 20\% & 1796.1 & $88.63\pm0.61$ & 26.76 \\
\bottomrule
\end{tabular*}
\end{table}

\begin{table}[!t]
\centering\small
\setlength{\tabcolsep}{1pt}
\caption{\textbf{Ablation of compression methods}. Log-mass MAE is the mean absolute error between predicted and dense block log-masses.}
\label{tab:tail-mechanism-ablation}
\begin{tabular*}{\columnwidth}{@{\extracolsep{\fill}}@{}>{\centering\arraybackslash}p{0.15\columnwidth}>{\centering\arraybackslash}p{0.16\columnwidth}>{\centering\arraybackslash}p{0.21\columnwidth}>{\centering\arraybackslash}p{0.21\columnwidth}>{\centering\arraybackslash}p{0.21\columnwidth}@{}}
\toprule
\textbf{Spark-Reblock} & \textbf{Spark-Reweight} & \textbf{DiT Latency (s) $\downarrow$} & \textbf{MAE $\downarrow$} & \textbf{PSNR (dB) $\uparrow$} \\
\midrule
 &  & 1765.0 & 3.38 & 21.24 \\
 & \checkmark & 1763.2 & 2.52 & 21.65 \\
\addlinespace[2pt]
\checkmark &  & 1796.0 & 1.39 & 26.24 \\
\checkmark & \checkmark & 1796.1 & 1.16 & 26.76 \\
\bottomrule
\end{tabular*}
\end{table}

\subsection{Compatibility with Few-Step Distillation}

\paragraph{Quantitative Results on MiniMax-H3-Turbo.} We further evaluate whether Spark-H3 can be combined with few-step diffusion distillation using MiniMax-H3-Turbo~\cite{lightx2v2026}, a community-released 8-step LoRA for MiniMax-H3. We generate 14.4-second videos at $1344\times768$ resolution using eight denoising steps, with two initial dense denoising steps as warm-up before switching to sparse attention.

As shown in Table~\ref{tab:lightx2v}, Spark-H3 achieves a 2.27$\times$ DiT speedup at 10\% exact-block density and 24.80 dB PSNR against Dense. Increasing the density to 20\% further improves PSNR to 26.48 dB while achieving a 2.08$\times$ speedup. At 20\% density, Spark-H3 also achieves higher PSNR and SSIM and lower LPIPS than all five baselines, while remaining faster than the three CSA baselines. These results demonstrate the effectiveness of Spark-H3 when combined with few-step distillation.

\subsection{Ablation studies}

\paragraph{Spark-Reblock and attention-mass recall.}
Table~\ref{tab:attn-recall} compares configurations with and without Spark-Reblock while keeping Spark-Reweight enabled. Attention-mass recall is the fraction of dense-softmax attention mass captured by selected blocks. Spark-Reblock increases recall by 18.33 and 14.58 percentage points at 10\% and 20\% density, respectively, and improves PSNR by 4.51 and 5.11 dB. Moreover, Spark-Reblock at $\rho=0.1$ achieves higher recall and PSNR with lower DiT latency than the configuration without Reblock at $\rho=0.2$, demonstrating a better quality--efficiency trade-off than doubling the exact-block budget. These results show that Spark-Reblock effectively addresses preference mismatch through preference-aware token grouping.

\begin{figure}[!t]
  \centering
  \includegraphics[width=\textwidth]{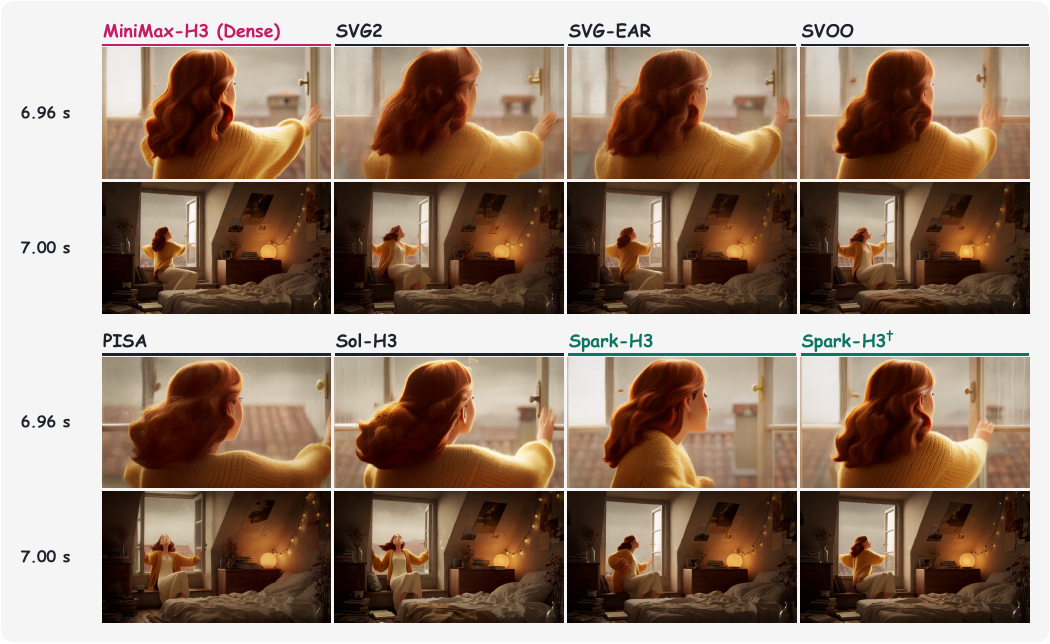}
  \caption{\textbf{Motion incoherency.}
  \textbf{Zoom in} to compare body poses between adjacent frames and hair details.}
  \label{fig:qual-attic}
\end{figure}

\paragraph{Spark-Reweight and compressed-block error.}
Table~\ref{tab:tail-mechanism-ablation} isolates Spark-Reweight at 20\% density with and without Spark-Reblock. Without Spark-Reblock, it reduces log-mass MAE from 3.38 to 2.52 and increases PSNR from 21.24 to 21.65 dB. With Spark-Reblock, it reduces log-mass MAE from 1.39 to 1.16 and increases PSNR from 26.24 to 26.76 dB. In both configurations, lower compressed-block mass error is accompanied by higher dense-reference fidelity. These results show that Spark-Reweight mitigates compression bias by reducing the Jensen gap in compressed-block attention-mass estimation.

\definecolor{hoodiecyan}{HTML}{53A4B9}
\definecolor{hoodieyellow}{HTML}{D2AA4B}
\begin{figure}[!t]
  \centering
  \includegraphics[width=\textwidth]{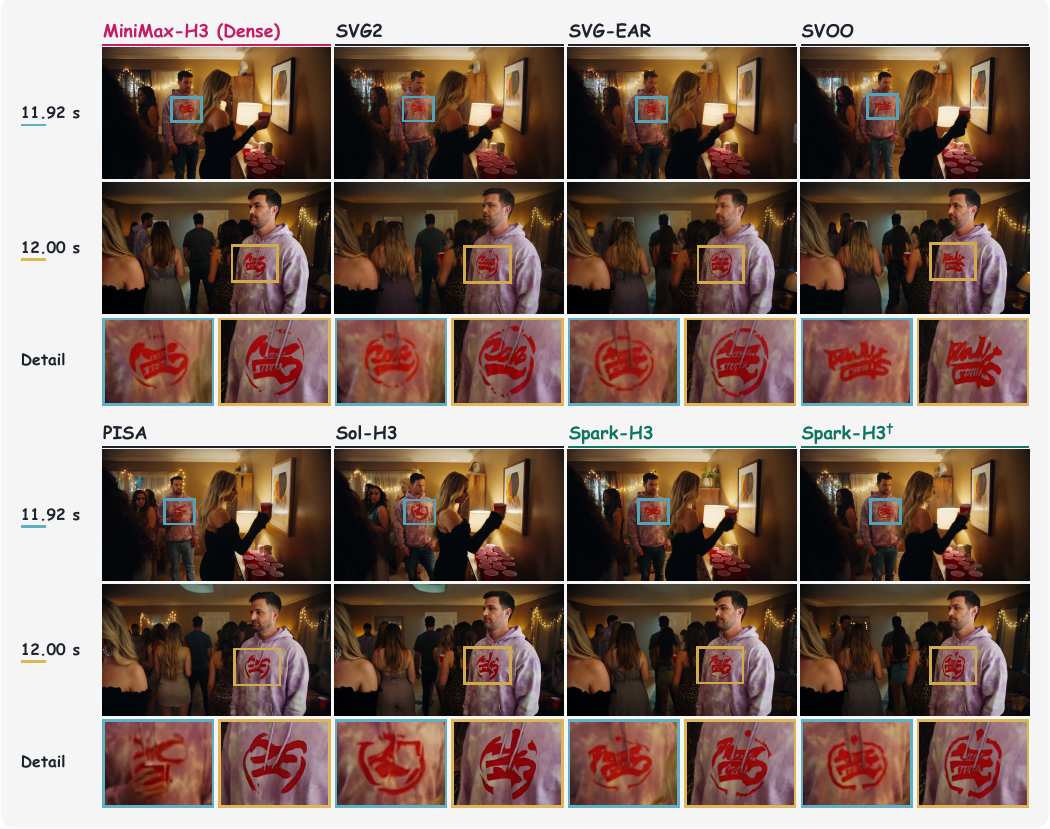}
  \caption{\textbf{Graphic inconsistency.}
  \textcolor{hoodiecyan}{Cyan} and \textcolor{hoodieyellow}{yellow} boxes mark the hoodie regions enlarged below; \textbf{zoom in} to compare the graphics.}
  \label{fig:qual-hoodie}
\end{figure}

\subsection{Qualitative results}
\label{sec:qualitative}

We present qualitative comparisons of MiniMax-H3 outputs for two prompts. In the attic scene (Fig.~\ref{fig:qual-attic}), Spark-H3 and SVG-EAR produce more plausible body poses and motion transitions between the two consecutive frames spanning the shot cut than the other sparse methods. Spark-H3 also retains finer hair detail than SVG-EAR. In the party scene (Fig.~\ref{fig:qual-hoodie}), the two BSA baselines, PISA and Sol-Attn, exhibit visible inconsistencies in the red hoodie graphic across the two views, whereas all three CSA methods and Spark-H3 preserve its structure more consistently. These examples illustrate Spark-H3's ability to preserve temporal coherence and fine visual details.

%% file: sec/spark_5_discussion.tex
\section{Conclusion}
We presented Spark-Attn, a training-free BSA method comprising Spark-Reblock and Spark-Reweight. Spark-Reblock uses quicksort-inspired recursive balanced splits to group tokens by attention preference into uniformly sized tiles, combining BSA's efficiency with CSA's adaptivity. Spark-Reweight reinterprets mean pooling as a first-order Taylor expansion at the zero query and shifts the expansion point to the global mean query, the center of the observed query distribution, yielding a new, principled, and lightweight compression approach. Extensive experiments across video diffusion models demonstrate Spark-Attn's effectiveness and generalizability, establishing it as a leading approach to block sparse attention optimization.